\documentclass[10pt,twocolumn,letterpaper]{article}

\usepackage{wacv}              

\usepackage[T1]{fontenc} 
\usepackage{multirow}
\usepackage{makecell}
\usepackage{pifont}
\usepackage{float}

\newcommand{\cmark}{\textcolor{green}{\ding{51}}}
\newcommand{\xmark}{\textcolor{red}{\ding{55}}}

\newcommand{\dsname}{AFSeq}

\usepackage{comment}

\definecolor{wacvblue}{rgb}{0.21,0.49,0.74}
\usepackage[pagebackref,breaklinks,colorlinks,allcolors=wacvblue]{hyperref}

\def\wacvPaperID{290} 
\def\confName{WACV}
\def\confYear{2027}

\title{From Next Day Wildfire to Next Days Wildfire: A New Multi-Continent Dataset of Active-Fire Spread for Sequence-to-Sequence Forecasting}

\author{Jose Guillen \\
CENIA \\
Macul, Chile \\
\small{\texttt{jose.guillen@cenia.cl}}
\and
Renzo Zanca \\
Universidad de Chile, DCC \\ 
Santiago, Chile \\
\small{\texttt{renzo.zanca@ug.uchile.cl}}
\and
Valentin Barriere \\
Universidad de Chile, DCC, CENIA \\ 
Santiago, Chile \\
\small{\texttt{vbarriere@dcc.uchile.cl}}
}

\begin{document}

\twocolumn[{
  \renewcommand\twocolumn[1][]{#1}
  \maketitle
  \begin{center}
    \captionsetup{type=figure}
    \includegraphics[width=\textwidth]{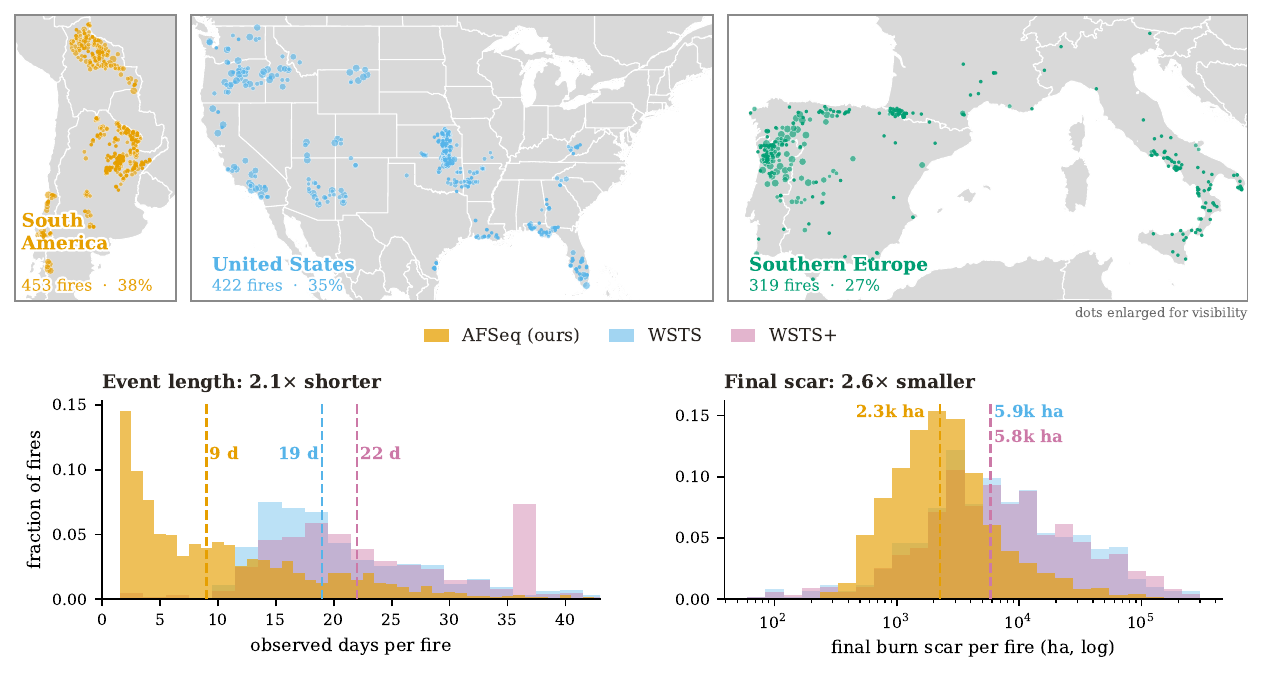}
    \captionof{figure}{\textbf{Overview of \dsname{}.} Each point denotes a single fire event. The dataset combines fire events from 8 countries over 3 regions: South America (Chile, Argentina, Bolivia), United States, and Europe (Portugal, France, Spain and Italy). Previous benchmarks \cite{gerard2023wildfirespreadts,lahrichi2026wstsplus} mainly focus on the United States and on mega-wildfires that mostly last more than 10 days. \dsname{} shifts the distribution to the left with shorter fires and smaller areas, which means \textbf{they are more common to see and harder to evaluate}.}
    \label{fig:teaser}
  \end{center}%
}]

\begin{abstract} 
Benchmarks such as WildfireSpreadTS (WSTS) have enabled data-driven forecasting of wildfire spread from multimodal satellite time series, but they suffer from limitations: size restricts them to large fires that burn for weeks, and they are from similar environments. 
We introduce \dsname{}, a multimodal dataset of 1,194 recent fire events (2023 to 2026) from 8 different countries across three regions, dominated by the short, small fires that existing benchmarks under-represent. Our dataset is the largest and most diverse to date for time-series-based wildfire spread prediction.
We also introduce UTAE-AR, an extension of a classical U-Net with Temporal Auto Encoder, enhanced with recurrent modules before the decoder. This allows the model to not only predict the next day but the whole sequence in an autoregressive way. Moreover, our model can feed its own predicted fire mask back one day at a time, producing the day-by-day forecasts that single-step models cannot.
Our results show that UTAE-AR surpasses state-of-the-art on both WSTS, WSTS+ and AFSeq for next-day accuracy while having 21.5\% of their parameters, and being able to predict confidently at a longer horizon.
\end{abstract}

\vspace{-1.3cm}
\section{Introduction}
\label{sec:intro}

Climate change is changing fire patterns around the world, and the cost to society is rising fast: the number of wildfires each year has more than quadrupled since 1980 \cite{cunningham2025disastrous}. 
These losses are expensive in the regions that get hit. In Southern Europe, an average fire season is linked to yearly production losses of EUR~13-21 billion~\cite{meier2023regional}. South America is on the same path: long droughts and higher temperatures lead to more frequent and severe fires~\cite{ciocca2023increased}. Controlling the 2017 megafires in Chile for example there cost more than USD~362 million in public money alone~\cite{gonzalez2020megafires}.

Wildfire spread prediction has been a task of interest for many years. Earlier work relied on physics-based simulation systems \cite{Finney_1998,rothermel1972}, which remain relevant but are computationally expensive and sensitive to parameter estimation. Machine Learning approaches are generally more robust and permit 
hybrid systems such as physics-informed neural networks \cite{hodges2019cnnfire,pereira2022review,dabrowski2023bpinn}. 
More recent approaches leverage multimodal images from satellite sensors \cite{huot2022next,Fitzgerald2023} to train CNN-type neural networks, and finally Satellite Image Time Series (SITS) \cite{gerard2023wildfirespreadts,Zhao2025a} to train recurrent or attention-based convolutional neural networks. 

From the WildfireSpreadTS (WSTS) benchmark~\cite{gerard2023wildfirespreadts}, two complementary branches of work have emerged: stronger encoders and more data for next-day prediction, culminating in the WSTS+ benchmark~\cite{lahrichi2026wstsplus}, with rigorous training and evaluation protocols for next-day prediction using single day and SITS inputs.
However, two gaps motivate our work. First, the distributions in WSTS and WSTS+ mainly focus on mega-wildfires with a median duration of 20 days. Smaller fires still matter, since every large fire is small in its early stages, when spread forecasts are most actionable. Second, although many works converge to the interest of using time-series \cite{gerard2023wildfirespreadts,Fitzgerald2023,li2024wildfire}, both modeling and evaluation have focused on next-day prediction; multi-day prediction, where the model uses its own predictions as input, is neither yet explored nor standardized.

\paragraph{Contributions of this Work.}
\label{par:contributions}

Our primary contribution is \dsname{}, a dataset of 1,194 multimodal daily active-fire sequences from three regions and several years (Europe, North America and South America; 2023-2026), covering the short, smaller, and recent fires that existing benchmarks under-represent. It is the most diverse dataset for wildfire propagation prediction as far as the authors know. Moreover, we show our dataset allows for better generalization in cross-dataset experiments. 

Our second contribution is at the algorithmic level. We propose UTAE-AR, an autoregressive extension of UTAE that is trained to predict multiple future fire masks sequentially. Rather than treating multi-day forecasting as the primary objective, this autoregressive supervision provides additional temporal structure during training through the use of a recurrent layer before the decoder that allows for sequence labeling inside of next-day wildfire prediction. Next-day prediction remains our primary target, since it is the main operational objective for firefighters. We show our new model improves the central task of next-day wildfire spread prediction on two datasets.

\section{Related Work}
\label{sec:related}

Wildfire spread forecasting has been studied through data-driven, physical, and hybrid approaches, with recent work increasingly exploiting multimodal satellite observations and temporal context. However, most learning-based methods and benchmarks remain centered on next-day prediction. We review the main modeling approaches and datasets relevant to this setting, with particular attention to the temporal horizon used for evaluation.

\paragraph{Spatial CNN approaches.} Early data-driven work predicts spread or final burned area with convolutional networks over satellite imagery, meteorology, and topography: FireCast~\cite{radke2019firecast} and Fire-Image-DenseNet~\cite{pang2025fidn} reach accuracy competitive with physical simulators at a fraction of the inference cost, and multi-scale, multi-kernel designs such as WFNet~\cite{jiang2020wfnet} and related CNNs~\cite{marjani2023multikernel,liu2023omscnn} improve the extraction of spatial context. These models score a single future state, one step ahead.

\paragraph{Temporal and attention models.} Sequence models improve on static inputs: ConvLSTM and 3D U-Net variants~\cite{kim2025assessment,anastasiou2025wildfire}, FirePred's multi-temporal processing~\cite{marjani2023firepred}, and CNN-BiLSTM hybrids~\cite{marjani2024cnn} exploit daily observation histories, and MA-Net~\cite{shadrin2024manet} adds positional and multi-scale attention over multimodal inputs. 
Recent works \cite{Fitzgerald2023, li2024wildfire, gerard2023wildfirespreadts, almakrami2025predicting} converge to the conclusion that attention-based U-Nets type of models allows for reaching higher performances. In particular UTAE~\cite{garnot2021utae}, combining convolutional encoding with temporal attention, sets the state of the art on WildfireSpreadTS~\cite{gerard2023wildfirespreadts}, and Lahrichi \etal~\cite{lahrichi2026wstsplus} report further gains from stronger encoders. Across this line of work, evaluation remains next-day.

\paragraph{SITS Datasets and benchmarks.} Next Day Wildfire Spread~\cite{huot2022next}, TS-SatFire~\cite{Zhao2025a}, WildfireSpreadTS~\cite{gerard2023wildfirespreadts}, and WSTS+~\cite{lahrichi2026wstsplus} pair fire masks with environmental and meteorological covariates, but mainly focus on large, long-growing fires that are uncommon in reality. Evaluation has centered on next-day prediction: none of these benchmarks standardizes multi-day evaluation and persistence baselines are not routinely reported. Another interesting dataset is FireSentry from \citet{Zhou2026}, which utilizes drones to gather sub-second and sub-meter resolution data, making it more valuable but at a higher collection price.

\section{The \dsname{} Dataset} 

Existing wildfire spread benchmarks are dominated by long and large fire events, leaving shorter and smaller fires comparatively under-represented. To complement this setting, we construct a new multimodal dataset with broader geographic coverage and an event-level design suited for multi-day forecasting. The following subsections describe its composition, construction pipeline, sampling strategy, data splits, and release.

\subsection{Overview}
\label{sec:afseq_overview}

We introduce \dsname{}, a dataset of multimodal daily active-fire sequences for multi-day spread forecasting. 
The final dataset has 1,194 fire events from 2023 to 2026, covering eight countries on three continents (\cref{fig:teaser}). \Cref{tab:populations} compares it with the two closest public benchmarks for this task working with SITS. 
\dsname{} offers advantages compared to the others. First, it is more diverse with respect to the geographical location of the data as it contains data from North America, Europe, and Latin America: this is as far as the authors know the only dataset for wildfire spread prediction using SITS of Latin America. Second, the dataset contains challenging events of rapid and small wildfires: the median observed days and final scar area is less than half of the other datasets. This makes the persistence baseline 35\% lower than for the other datasets. 
Additional descriptive statistics including event counts by region, country, and year, as well as per-channel visualizations and breakdowns by land cover and vegetation indices are provided in the supplementary material (see \cref{app:sec:data_details}).

\begin{table}[t]
\centering
\caption{\dsname{} compared with the two public SITS benchmarks.}
\label{tab:populations}
\small
\setlength{\tabcolsep}{4pt}
\resizebox{\linewidth}{!}{
\begin{tabular}{lccc}
\toprule
 & \dsname{} & WSTS \cite{gerard2023wildfirespreadts} & WSTS+ \cite{lahrichi2026wstsplus} \\
\midrule
\multicolumn{4}{l}{\emph{Coverage}} \\
Years covered            & 2023-2026 & 2018-2021 & 2016-2023 \\
Regions                  & \makecell{US, S. America,\\S. Europe} & Western US & US \\
\midrule
\multicolumn{4}{l}{\emph{Size}} \\
Fire events              & $1,194$ & 607 & $1,005$ \\
Total daily obs.         & $13,842$ & $13,607$ & $24,462$ \\
\midrule
\multicolumn{4}{l}{\emph{Population}} \\
Median observed days     & \textbf{9} & 19 & 22 \\
Median final scar (ha)   & \textbf{2.3k} & 5.9k & 5.8k \\
\midrule
\multicolumn{4}{l}{\emph{Difficulty baseline}} \\
Persistence AP & $\mathbf{0.124 \pm 0.013}$ & $0.193 \pm 0.071$ & $0.200 \pm 0.063$ \\
\bottomrule \vspace*{-.8cm}
\end{tabular} 
}
\end{table}

\subsection{Construction}
\label{sec:afseq_construction}

Our dataset is built from public satellite and environmental products using Google Earth Engine. The building pipeline consists of four stages: candidate discovery, per-day extraction, curation and sampling.

\paragraph{Candidate discovery.}
We start by looking for groups of burned pixels that are connected to each other, using the MODIS MCD64A1 v061 burned-area product. We keep only groups that are larger than 300 ha, have less than 50\% cropland, burn on at least two different days and overlap any already collected event by less than 30\% of the smaller event's area.

\paragraph{Per-day extraction.}
Each event gets its own 200-meter grid. This grid is a window that covers the event's final burned area, plus a 10 km margin on all sides. The window is never smaller than 256x256 pixels (51 km). 
For each observed day, we pull:
\begin{itemize}
    \item Daily active-fire mask from VIIRS combining Suomi-NPP and NOAA-20, both are same instrument and pass 50 minutes apart.
    \item Three VIIRS surface reflectance bands (I1, I2 and M11) from VNP09GA.
    \item Seven daily weather variables: surface temperature, dew point, relative humidity, wind speed and direction, surface pressures from MODIS and ERA5 and precipitation from CHIRPS.
    \item Elevation, slope and aspect from SRTM.
    \item Landcover, NDVI, EVI and evapotranspiration from MODIS.
\end{itemize}

Missing data are stored as NaN. Later when loading the data, we fill in missing values with the average of that channel from the training set. About 3.4\% of the reflectance days have missing data.

Weather is the one family where we depart from WSTS\cite{lahrichi2026wstsplus,gerard2023wildfirespreadts}, because GRIDMET variables (ERC, PDSI) exist only for United States, so we use their global counterparts at the cost of coarser resolution.

\paragraph{Sampling.}

Each sample consists of a day $t$ when the fire is actively burning, together with the four preceding days (five observed frames in total, corresponding to k = 4 in \cref{sec:task}) and the following days as target outputs. 
The models do not see the full temporal window at once. Instead, they are trained and evaluated on \(128 \times 128\) pixel patches (\(25.6\,\text{km}\)), matching the crop size used in WSTS+\cite{lahrichi2026wstsplus}. Each patch is centered around the active fire of day \(t\) but with a random shift of up to a quarter of the patch size, ensuring that the crop is genuinely random and that the window border never appears inside it. This prevents the model from being biased toward an active fire at the center of the patch. 
Since 96.4\% of the events are smaller than the training patch, the task is fair in relation to the burned area: the model consistently sees the entire active fire mask without cuts at the edges.

\begin{figure*}
    \centering \vspace*{-.3cm}
    \includegraphics[width=\linewidth]{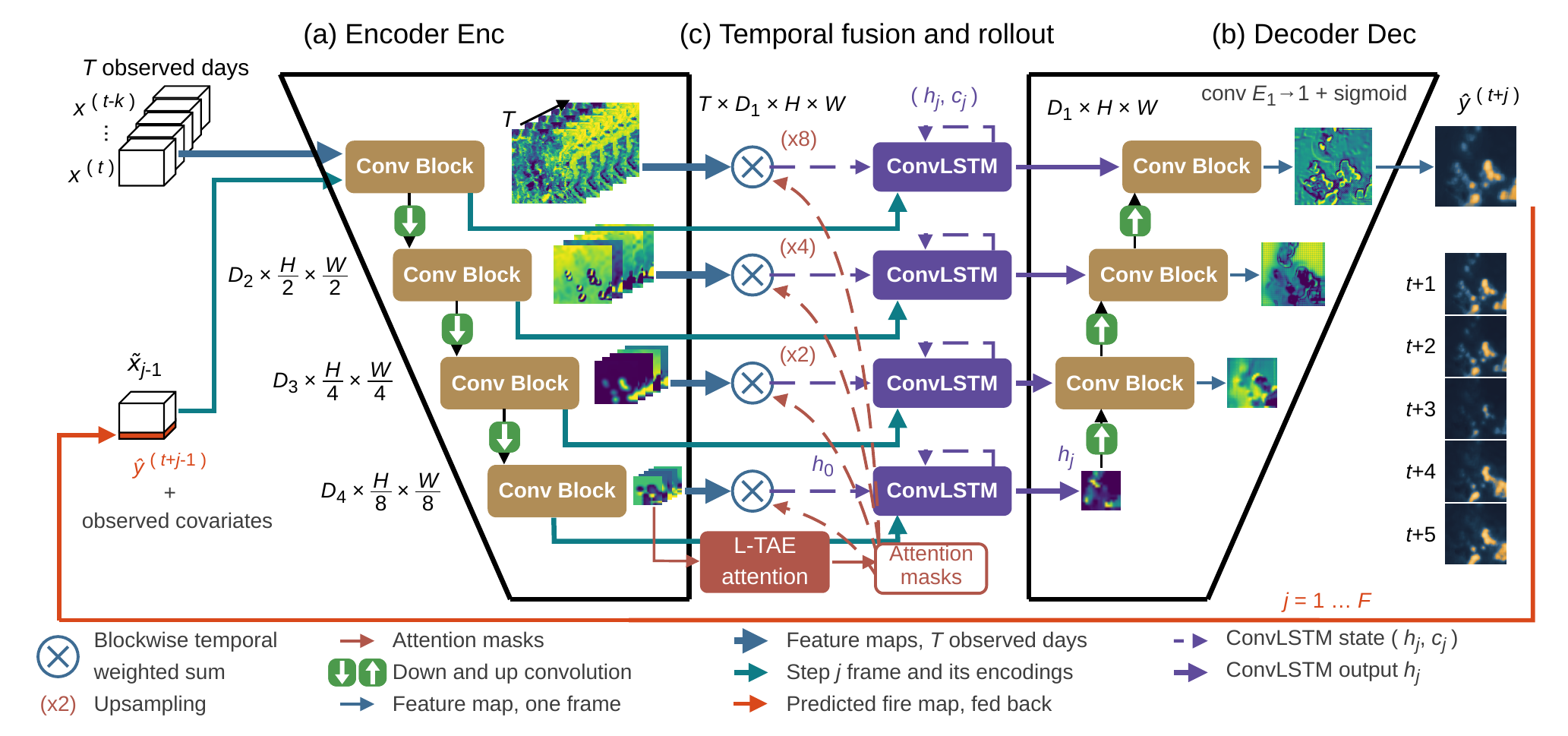} \vspace*{-.5cm}
    \caption{Overview of UTAE-AR architecture.} \vspace*{-.5cm}
    \label{fig:utae-ar}
\end{figure*}

\paragraph{Splits and leakage.}

Events are split 70/15/15 by fire identity so that no observation from the same event appears in more than one partition: 835 training, 179 validation, and 180 test events. We balance the partitions by both region and fire density (the mean fraction of burning pixels per-event). We create three partitions with fixed seeds and release all of them. Additional details are provided in the supplementary material (see \cref{app:sec:data_details}).

\subsection{Availability and License}
\label{sec:afseq_release}

The dataset is available at \href{https://dataverse.harvard.edu/previewurl.xhtml?token=be6fd344-7b1f-46c8-9873-84a8be64f3d2}{Harvard Dataverse} under \textbf{CC BY 4.0}, with construction and training code at \href{https://github.com/joseg20/AFSeq}{Github Repository}. All upstream products are open data that permit redistribution; ERA5-Land requires attribution to the Copernicus Climate Change Service, which we provide. Data were retrieved through Google Earth Engine under its non-commercial terms.

\section{Training and Validation Methodology}
\label{sec:methodology}

We formulate wildfire spread prediction as a multi-day forecasting problem and define a common evaluation protocol across datasets and models. The methodology is designed to handle variable event durations while avoiding information that would not be available at deployment time. We first define the forecasting task and its temporal constraints and then describe the metrics and benchmarks evaluation settings used throughout the experiments.

\subsection{Task}
\label{sec:task}

Let $x^{(t)} \in \mathbb{R}^{C \times H \times W}$ be the multimodal observation of an event on day $t$, with $C$ input channels over an $H \times W$ patch, and $y^{(t)} \in \{0,1\}^{H \times W}$ the corresponding active-fire mask. The model observes the past $k+1$ days, $x^{(t-k)}, \dots, x^{(t)}$, and predicts the fire masks of the next $F$ days, $y^{(t+1)}, \dots, y^{(t+F)}$. We use $k=4$ and $F=5$. All sequences were padded to a length $H$ to keep the short ones. 

\paragraph{Covariate availability.} 
The model never sees data from a day after t. For next-day prediction it uses only data up to day $t$. For the multi-day, each step uses a weather forecast made on day $t$ for that future day. This matches what a user would have in practice: the real weather of the coming days is not known yet, only a forecast of it.

\subsection{Models}

We compare standard single-frame and temporal baselines alongside our proposed autoregressive model, UTAE-AR. While the baselines differ in the amount of temporal context that they use, all are designed for next-day prediction. UTAE-AR extends this setting to multi-day forecasting.

\subsubsection{UTAE-AR}

UTAE-AR extends UTAE with an autoregressive prediction mechanism. It keeps the multi-scale encoder and the L-TAE temporal attention unchanged, and introduces the recurrence \emph{at the temporal-fusion stage}. In UTAE-AR, the attention-weighted temporal sum at each U-Net level initializes the hidden state of a ConvLSTM cell placed at that level. Fig.~\ref{fig:utae-ar} shows the architecture.

\paragraph{Rollout.}
Let $x^{(t-k)},\ldots,x^{t}$ denote the observed history of $T$ days, $\mathrm{Enc}_\ell$ the encoder output at level $\ell=1,\ldots,4$ and $h^{\ell}_j,c^{\ell}_j$ the state of the ConvLSTM cell at level $\ell$ after rollout step $j$. The history is encoded once and fused by the L-TAE which initializes the cell states,

\begin{equation}
    h^{(\ell)}_0 = \mathrm{LTAE}_\ell\!\left(x^{(t-k)},\ldots,x^{(t)}\right),
    \qquad
    c^{(\ell)}_0 = 0.
    \label{eq:ar_init}
\end{equation}

The rollout then proceeds one forecast day at a time. Let $\tilde{x}_{j-1}$
denote the input frame at step $j$: for $j=1$ it is the last observed frame
$x^{(t)}$, and for $j\geq 2$ it combines the observed covariates of day
$t+j-1$ with the fire mask fed back for that day. Each step consists of
three operations, applied identically at each of the four encoder levels.
First, the encoder processes the single input frame,

\begin{equation}
    e_j = \mathrm{Enc}(\tilde{x}_{j-1}).
\end{equation}

Second, these features update the ConvLSTM state,

\begin{equation}
    h_j = \mathrm{ConvLSTM}(e_j, h_{j-1}),
\end{equation}

so the state carries the fire history forward.
Third, the decoder reads the updated states of the four levels and
produces the next mask,

\begin{equation}
    \hat{y}^{(t+j)} = \sigma\big(\mathrm{Dec}(h_j)\big),
\end{equation}

where $\mathrm{Dec}$ includes the output head and $\sigma$ is the sigmoid. 

Following WSTS+\cite{lahrichi2026wstsplus}, our temporal encoder uses the position within the observed window, $\bar{t}\in\{1,\ldots,T\}$, as the positional index, instead of the day of year used in the original UTAE~\cite{garnot2021utae}. Additional architectural details are provided in the supplementary material (see \cref{app:sec:utae_det}).

\paragraph{Autoregressive training.}
Teacher forcing consists in using the ground truth as input of the autoregressive model when training. Using it alone leaves the model unprepared for its own errors at inference (i.e., exposure bias~\cite{bengio2015scheduled}), so we use scheduled sampling: at step $j\geq 2$, the mask fed back is the ground truth $y^{(t+j-1)}$ with probability $p_e$ and the prediction $\hat{y}^{(t+j-1)}$ otherwise, drawn independently per sample and step. We decay $p_e$ linearly from $1$ to a floor of $0.5$ over training, $p_e=\max\!\big(0.5,\,1-\tfrac{e}{2(E-1)}\big)$, where $e$ is the current epoch and $E$ the total. $p_e=1$ means full teacher forcing while $p_e=0$ only using its own predictions. 

In \cref{sec:ablation} we compare this linear decay against free running ($p_e=0$ throughout) and against inverse-sigmoid scheduled sampling~\cite{bengio2015scheduled}, $p_e=\kappa/\big(\kappa+\exp(e/\kappa)\big)$, where $\kappa$ sets how long training is held close to teacher forcing: $\kappa{=}1$ is near free running from the start, while $\kappa{=}\infty$ keep $p_e \approx 1$ across all epochs.

Note that the rollout conditions on the observed covariates of each future day, so it forecasts fire spread given the weather rather than the weather itself.

\subsubsection{Baselines}

We evaluate two families. \emph{Single-day input} models see only day $t$: a U-Net \cite{ronneberger2015unet} and a ConvNeXt based segmentation network \cite{liu2022convnext}. \emph{Time Series input} models consume the last $k+1$ days of history: ConvLSTM \cite{shi2015convlstm}, UTAE \cite{garnot2021utae}, which sets the state of the art on WSTS with a temporal attention encoder, and a Res18-UTAE, the strongest configuration reported on WSTS+ \cite{lahrichi2026wstsplus}. All predict a single $t+1$ future mask.

\subsection{Metrics}

We report average precision (AP) on the predicted probability maps at $t+1$, consistently with WSTS~\cite{gerard2023wildfirespreadts} and WSTS+~\cite{lahrichi2026wstsplus}. 
In this way, although UTAE-AR produces multi-day forecasts, autoregressive supervision acts as additional temporal structure during training and improves next-day performance. 

On \dsname{} every model is trained and scored with all channels of the dataset. On WSTS+ we follow its published protocol exactly: the four two-year folds, which contain the WSTS+\cite{lahrichi2026wstsplus} additions together with the original WSTS\cite{gerard2023wildfirespreadts} years, the same three feature sets (Veg, Multi, All) and the same protocol of training and evaluation. 
Experimental details and key deployment information are provided in the supplementary material (see \cref{app:sec:exp_det}).

\section{Experiments and Results}
\label{sec:protocol}

The experiments evaluate UTAE-AR from three complementary perspectives: in-domain next-day prediction, transfer across wildfire benchmarks, and the effect of autoregressive supervision over increasing forecast horizons. Results are reported on \dsname{} and WSTS+ under their respective evaluation protocols, followed by cross-dataset transfer and an ablation over the supervision of UTAE-AR.

\subsection{Experimental setup}
\label{sec:exp_setup}

Our models are trained with the same protocol: focal loss, Adam at $10^{-3}$, and checkpoint selection by validation AP under free-running rollout.
On WSTS+ we follow the four-fold biannual protocol of \cite{lahrichi2026wstsplus} and report the mean and standard deviation across folds. 
On \dsname{} we train and evaluate on a 70/15/15 stratifield-k-fold splits, operation we repeat three times on three different splits and report the mean and standard deviation across them. 
Res18-UTAE and UTAE follows the protocol of \cite{lahrichi2026wstsplus}, with relative temporal position encoding and an extensive hyperparameter search comprising more losses and learning rates. 

\subsection{Results on AFSeq}
\label{sec:res_afseq}

We first evaluate next-day prediction on \dsname{}, comparing persistence, single-frame, and temporal models under the protocol of \cref{sec:exp_setup}. Results are reported in \cref{tab:afseq_panel}.

\begin{table}[t]
\centering
\caption{Next-day prediction on \dsname{}. AP on the test partition, mean and standard deviation over three seeds. Single-frame models see day $t$ only; temporal models see the five-day history.}
\label{tab:afseq_panel}
\small
\setlength{\tabcolsep}{4pt}
\resizebox{.9\linewidth}{!}{%
\begin{tabular}{lccrc}
\toprule
Model & Seq in & Seq out & Params & AP \\
\midrule
Persistence            & \xmark & \xmark & $\emptyset\quad$      & $.124 \pm .013$ \\
\midrule
\multicolumn{5}{l}{\emph{Single-frame}} \\
U-Net                  & \xmark & \xmark & 0.43 M  & $.242 \pm .032$ \\
ConvNeXt               & \xmark & \xmark & 203 M   & $.254 \pm .037$ \\
\midrule
\multicolumn{5}{l}{\emph{Temporal}} \\
ConvLSTM               & \cmark & \xmark & 0.19 M  & $.302 \pm .028$ \\
UTAE                   & \cmark & \xmark & 0.89 M  & $.298 \pm .035$ \\
Res18-UTAE          & \cmark & \xmark & 14.7 M  & $.321 \pm .013$ \\
\textbf{UTAE-AR}       & \cmark & \cmark & 3.16 M  & $\mathbf{.382 \pm .022}$ \\
\bottomrule
\end{tabular}} \vspace*{-.3cm}
\end{table}

Models that use temporal context consistently outperform the single-frame baselines. The single-frame models obtain similar performance, whereas the temporal baselines benefit from the five-day observation history. 

Res18-UTAE provides the strongest non-autoregressive baseline in WSTS+\cite{lahrichi2026wstsplus}. UTAE-AR improves over the strongest non-autoregressive baseline by \textbf{19\%} relative AP, \textbf{while using 78.5\% fewer parameters}. Relative to persistence, UTAE-AR improves AP by \textbf{208\%}, indicating that next-day spread cannot be captured simply by carrying the current fire state forward.

Beyond model performance, \dsname{} places greater emphasis on shorter and smaller fire events than existing spread benchmarks. Using the fire-size classification of Engle \etal~\cite{engle2025boltfire}, small fires are those up to 1,000 ha and moderate fires range from 1,000 to 10,000 ha. These size classes correspond to the more frequently occurring portion of observed fire-size distributions. In the BoLtFire dataset, small and moderate fires together account for 90.37\% of fire events \cite{engle2025boltfire}. A recent large-scale analysis of Eurasian drylands similarly found that fires below 1,000\,ha accounted for 76.71\% of events, whereas fires above 10,000 ha represented only 2.2\% \cite{yu2026fuel}. By shifting the benchmark toward these smaller events, \dsname{} complements existing datasets that emphasize larger and longer-lasting fires. A complete breakdown of \dsname{}, WSTS, and WSTS+ by fire-size class is provided in the supplementary material (see \cref{app:sec:supp_results}).

\subsection{Results on WSTS+}
\label{sec:res_wstsplus}

On WSTS+ we follow the published four-fold splits of \citet{lahrichi2026wstsplus}, defined over all eight years of the benchmark. The years are grouped into four blocks of two consecutive years. In each fold, one block is used for testing, one for validation, and the remaining two for training. Every block is used once for testing and once for validation. This splitting was introduced to reduce the computational cost of the original twelve-fold WSTS evaluation while preserving temporal separation between training and evaluation years. 
The original WSTS splitting is defined over its four years and is therefore not directly comparable with the WSTS+ protocol. The random four-fold split over fire events also reported by \citet{lahrichi2026wstsplus} corresponds to a different evaluation setting. We therefore compare results only within the same protocol. Results on WSTS under its original twelve-fold protocol are reported in the supplementary material (see \cref{app:sec:supp_results}).

\begin{table}[t]
\centering
\caption{Next-day prediction on WSTS+ under its published four-fold protocol. AP on the 128-pixel crop, reported as mean and standard deviation across folds.}
\label{tab:wstsplus}
\small
\setlength{\tabcolsep}{4pt}
\resizebox{\linewidth}{!}{%
\begin{tabular}{lcccc}
\toprule
Model & Veg & Multi & All & Params\\
\midrule
Res50-Unet \cite{lahrichi2026wstsplus} & $.349 \pm .109$ & $.351 \pm .105$ & $.351 \pm .112$ & 32.5 M \\
\midrule
Res18-UTAE \cite{lahrichi2026wstsplus} & $.354 \pm .113$ & $.363 \pm .129$ & $.350 \pm .117$ & 14.7 M \\
\midrule
UTAE-AR & $\mathbf{.429 \pm .105}$ & $\mathbf{.445 \pm .103}$ & $\mathbf{.441 \pm .106}$ & \textbf{3.16 M}\\
\bottomrule
\end{tabular}} \vspace*{-.2cm}
\end{table}

\Cref{tab:wstsplus} compares UTAE-AR with the strongest published WSTS+ baselines under the same evaluation protocol. Despite using a fixed learning rate of $10^{-3}$ without grid search or pretrained initialization, UTAE-AR achieves the highest AP across all feature sets. Relative to the best published baseline in each setting, the improvement ranges from \textbf{21.2\%} to \textbf{25.6\%}, with an average improvement of \textbf{23.1\%}. UTAE-AR also uses \textbf{78.5\%} fewer parameters than Res18-UTAE, showing that the gain does not come from increased model capacity.

\section{Models and Results Analysis}
\label{sec:analysis}

\subsection{Transfer between benchmarks}
\label{sec:cross}
We next examine whether models trained on one benchmark generalize to the other. To make the comparison possible, all transfer experiments use the \emph{Veg} feature space shared by \dsname{} and WSTS+. The weather channels of WSTS+ come from specific products from United States. We evaluate each model both in domain and across datasets, and additionally consider joint training on the two benchmarks.

\begin{table}[t]
\centering
\caption{Cross-dataset evaluation: AP at $t{+}1$. Rows indicate the training set used; columns indicate the test set. WSTS+ columns correspond to the mean of the 4 folds. AFSeq columns correspond to the mean of the 3 splits. The row labeled \emph{Both} trains on the union of the two training sets of \emph{Veg}.}
\label{tab:transfer}
\small
\setlength{\tabcolsep}{4pt}
\resizebox{\linewidth}{!}{%
\begin{tabular}{llcccc}
\toprule
Model & Trained on & \dsname{} & WSTS+  & Avg.\\
\midrule
\multirow{3}{*}{UTAE-AR}  & WSTS+     & $.238 \pm .027$ & $\mathbf{.429 \pm .105}$ & .334\\
 & \dsname{} & $\mathbf{.354 \pm .140}$ & $.343 \pm .096$ & .349  \\
 & Both      & $.346 \pm .010$ & $.416 \pm .118$ & .381\\
\midrule
\multirow{3}{*}{Res18-UTAE}  & WSTS+     & $.234 \pm .012$ & $.354 \pm .113$ & .294\\
 & \dsname{} & $.317 \pm .012$ & $.305 \pm .087$ & .311 \\
 & Both      & $\mathbf{.331 \pm .07}$ & $\mathbf{.406 \pm .111}$ & .369\\
\bottomrule
\end{tabular}} \vspace*{-.2cm}
\end{table}

\Cref{tab:transfer} shows a clear performance drop when models are trained on one benchmark and evaluated on the other. The transfer is also asymmetric. For UTAE-AR, training on \dsname{} reaches $0.343$ AP on WSTS+, compared with $0.429$ when trained directly on WSTS+, while training on WSTS+ reaches only $0.238$ AP on \dsname{}, compared with $0.354$ when trained directly on it. A similar pattern appears for Res18-UTAE. This suggests that the two benchmarks capture different fire populations and that performance learned on one does not transfer equally to the other.

Training on both datasets largely closes this gap. The goal of joint training is not to obtain the best in-domain score on each test set, but a single model that remains robust across both benchmarks. UTAE-AR reaches $0.346$ AP on \dsname{} and $0.416$ on WSTS+, about one AP point below its performance when trained separately on each benchmark, but with the best average across both test sets. Res18-UTAE benefits even more from joint training, improving over its single-dataset results on both benchmarks. These results indicate that the two datasets provide complementary training examples rather than redundant information.

The combined training set is nevertheless dominated by United States events, because WSTS+ is entirely focused on that country. We do not consider this a flaw of the joint training setup itself, but a consequence of the current benchmark landscape, where most available wildfire spread data come from a \emph{single geographic domain}. The joint setting should therefore not be read as a geographically balanced benchmark, and the imbalance further illustrates the need for broader regional coverage. A more balanced combination of events across regions may improve transfer outside the United States.

\subsection{Autoregressive supervision and forecast horizon}
\label{sec:ablation}

The previous experiments focus on next-day prediction. We now extend the evaluation beyond $t+1$ and examine how UTAE-AR behaves as it rolls forward through $t+2,\ldots,t+F$. We also compare different autoregressive supervision strategies to study how the training procedure affects performance as the forecast horizon increases. 

For the multi-day analysis, we reuse the strongest temporal baseline trained for sequence to next-day prediction and roll it forward recursively as a reference against UTAE-AR.

This comparison allows us to distinguish between simply applying a next-day model repeatedly and explicitly training the model with autoregressive supervision. We then evaluate both approaches from $t+1$ through $t+H$.

\begin{figure}[t]
    \centering
    \includegraphics[width=\linewidth]{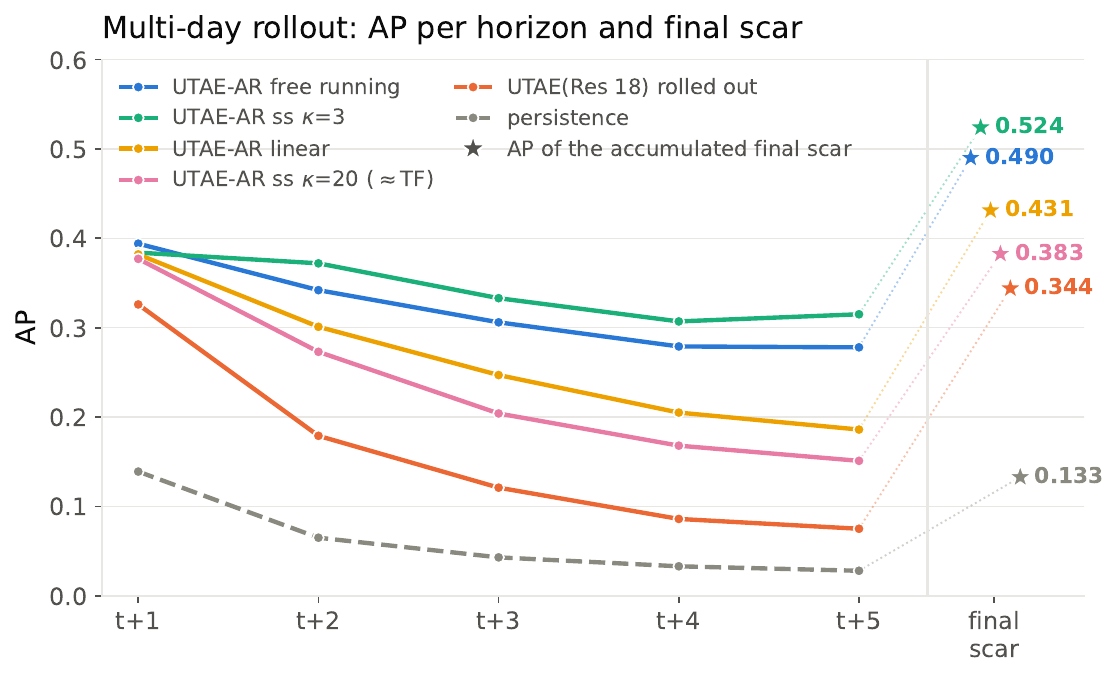}
    \caption{Multi-day forecast performance across horizons of UTAE-AR.}
    \label{fig:autoregressive_ablation} \vspace*{-.2cm}
\end{figure}

\Cref{fig:autoregressive_ablation} shows that all methods perform worse as the forecast horizon increases, but the rate of degradation differs substantially. The Res18-UTAE baseline performs competitively at $t+1$ but degrades rapidly when rolled out recursively, showing that strong next-day performance does not necessarily translate into stable multi-day forecasts. For the UTAE-AR versions, the free-running approach does best at $t+1$. However, scheduled sampling with $\kappa=3$ performs best starting at $t+2$ and maintains the highest average precision through $t+5$.

The same trend appears in the accumulated final scar. Scheduled sampling with $\kappa=3$ reaches $0.524$ AP, compared with $0.490$ for free running and $0.344$ for the recursively rolled Res18-UTAE, while persistence reaches only $0.133$. These results show that autoregressive supervision becomes more important as predictions move further from the observed history, and that scheduled sampling provides a better balance between next-day accuracy and stability across future horizons.

\subsection{Domain shift in representation space}
\label{sec:representation}

To examine how the two benchmarks differ beyond predictive performance, we compare the feature space learned by the encoder when trained on WSTS+ alone and on both datasets.

\begin{figure}[t]
    \centering
    \includegraphics[width=\linewidth]{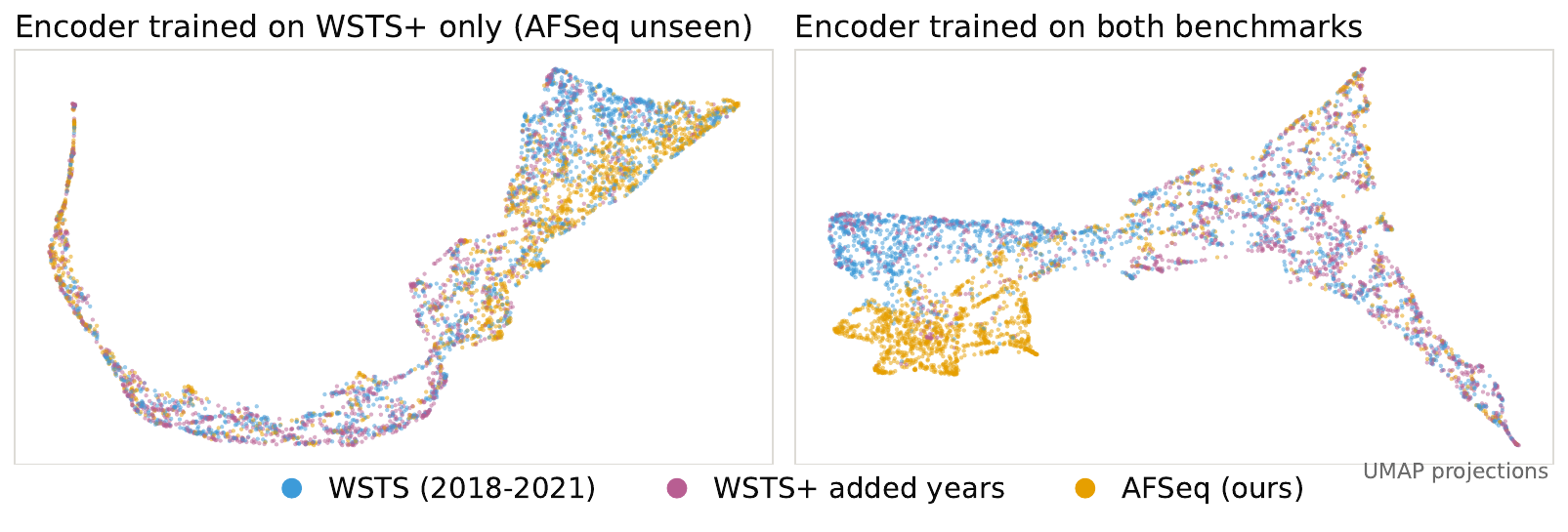}
    \caption{Encoder representations across benchmarks. Joint UMAP of the 512-dimensional deepest-stage features. With WSTS+ training only, unseen \dsname{} samples mostly fall within WSTS+ neighborhoods. Joint training reveals a distinct \dsname{} mode while preserving substantial overlap.}
    \label{fig:umap}
\end{figure}

\Cref{fig:umap} shows that an encoder trained only on WSTS+ represents most unseen \dsname{} samples within neighborhoods already occupied by WSTS+, with only 5\% having exclusively \dsname{} neighbors. This is not explained by the inputs alone, where the same analysis separates 34\% of \dsname{} samples. After training on both benchmarks, this fraction rises to 44\%, while 25\% of \dsname{} samples still lie in WSTS+ majority neighborhoods. The combination of overlap and separation suggests that \dsname{} contains both fire regimes shared with WSTS+ and regimes that are not well represented by the encoder trained just on WSTS+.

\subsection{Qualitative Results on \dsname{}}
\label{sec:q_results}

\begin{figure}[t]
    \centering
    \includegraphics[width=\linewidth]{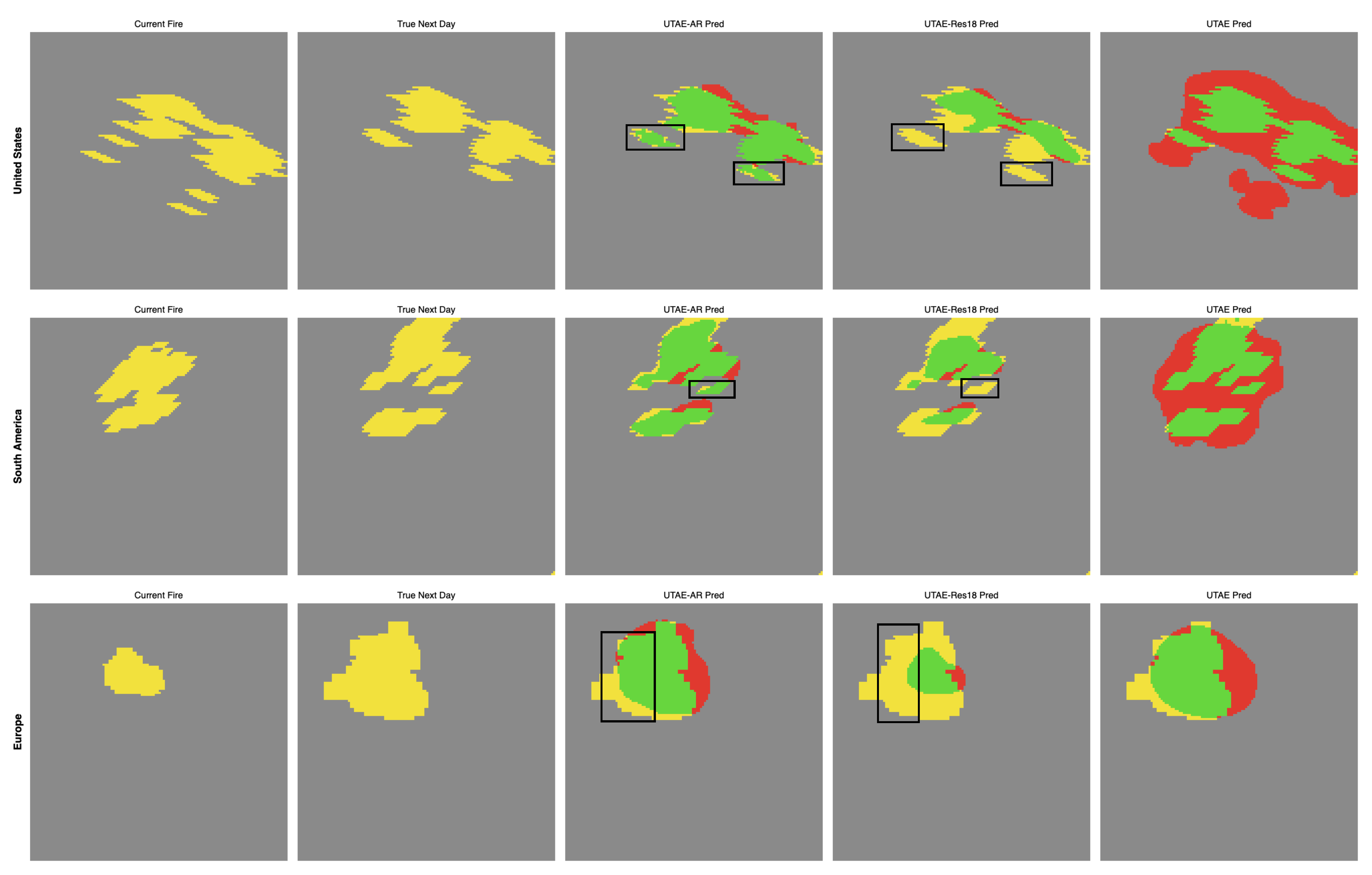}
    \caption{Qualitative next-day predictions on \dsname{} from the United
    States, South America, and Europe. Green, red, and yellow denote true
    positives, false positives, and false negatives, respectively. Black boxes
    highlight regions where UTAE-AR better recovers the observed fire spread
    than Res18-UTAE.} \vspace*{-.3cm}
    \label{fig:qr}
\end{figure}

We complement the quantitative evaluation with qualitative next-day predictions from the three regions covered by \dsname{}. 
Additional examples of cases where UTAE-AR performs better and worse are provided in the supplementary material (see \cref{sec:add_results}).

\section{Limitations and Future Work}
\label{sec:limitations}

\paragraph{Performance remains sensitive to fire size.} Prediction quality decreases for smaller fire regions, as shown by the per event analysis in the supplementary material (see \cref{app:sec:supp_results}). This is particularly relevant for \dsname{}, which contains a larger proportion of short and small fires than existing benchmarks. These events provide fewer positive pixels and less developed spatial structure, making them harder to predict. Yet, relative to persistence, the models retain more skill at lower burned areas on \dsname{} than on the existing benchmarks (see \cref{app:sec:data_details}). This suggests that the benchmark does not simply make the task harder, but also encourages models to learn useful signals in a regime that is under-represented in current datasets. Future work should further improve performance for the smallest and earliest fire stages.

\paragraph{Geographic generalization is still limited.} Although \dsname{} extends wildfire-spread evaluation beyond the predominantly US-centered setting of existing benchmarks, it does not yet constitute a global benchmark. Expanding \dsname{} to additional regions would extend the regional generalization analysis reported in the supplementary material (see \cref{app:sec:loro}) and enable a more systematic evaluation across geographic settings.

\paragraph{Multi-day modeling remains open.} UTAE-AR shows that supervision over multiple future days can improve next-day prediction, even when $t+1$ remains the primary evaluation target. However, autoregressive rollout is only one way to introduce future temporal structure. There is substantial room for architectures that model several future states jointly, reduce error propagation across horizons, or learn more explicit representations of fire dynamics. Extending existing next-day approaches to multi-day prediction would also enable broader comparisons between temporal modeling strategies under a common evaluation setting.

\section{Conclusion}
\label{sec:conclusion}

The event population used for evaluation determines which aspects of wildfire spread a benchmark measures. Existing benchmarks have enabled substantial progress in next-day prediction, but their event distributions place greater weight on relatively large and long fires and remain geographically concentrated in the United States. \dsname{} complements this setting by shifting evaluation toward shorter, smaller and moderate events while extending coverage across multiple regions. On top of that, we show with UTAE-AR that modeling the spread prediction task as sequence labeling allows to greatly improve the performance of the model. 

We introduce \dsname{}, a dataset of 1,194 multimodal daily active-fire sequences from eight countries across three regions between 2023 and 2026. Its broader event population provides a common setting for studying wildfire spread beyond the predominantly US-centered benchmarks currently available. We also introduce UTAE-AR, which uses autoregressive supervision over future fire states while retaining next-day prediction as the main objective.

On \dsname{}, UTAE-AR improves over the strongest non-autoregressive baseline by 19.0\% relative AP while using 78.5\% fewer parameters. On WSTS+, it outperforms the strongest published baseline across all evaluated feature sets, with relative AP improvements between 21.2\% and 25.6\%. Together with the cross-benchmark results, these findings show that both the composition of the fire population and the temporal formulation of the task have a substantial effect on measured wildfire spread performance.

By releasing \dsname{} and UTAE-AR, we aim to broaden wildfire spread research toward more representative fire populations, wider geographic coverage, and models that reason about fire evolution beyond a single future day.


{
    \small
    \bibliographystyle{ieeenat_fullname}
    \bibliography{main,Wildfire}
}

\newpage
\setcounter{page}{1}
\maketitlesupplementary

\renewcommand{\thesection}{S\arabic{section}}
\renewcommand{\thetable}{S\arabic{table}}
\renewcommand{\thefigure}{S\arabic{figure}}

\section{\dsname{} Data Details}
\label{app:sec:data_details}

\subsection{Channels}
\label{app:sec:channels}

Every event carries 34 input channels on a common 200 m grid: three static terrain channels, land cover as 17 one-hot IGBP classes, three vegetation indices, seven daily weather channels, three daily VIIRS reflectance bands, and the daily active-fire mask, which is both the last input channel and the prediction target. \cref{tab:channels} lists their products. Sources are never back-filled with a plausible constant: a missing observation is written as NaN, and imputed with the channel mean by the loader, so a hole always remains distinguishable from a measurement.

\begin{table*}[t]
\centering
\setlength{\tabcolsep}{4pt}
\small
\begin{tabular}{llll}
\toprule
Channel(s) & Product & Native res. & Cadence \\
\midrule
elevation, slope, aspect & SRTM & 30\,m & static \\
landcover (17 one-hot) & MODIS MCD12Q1 (IGBP) & 500\,m & annual \\
ndvi, evi & MODIS MOD13Q1 & 250\,m & 16\,d \\
et & MODIS MOD16A2 & 500\,m & 8\,d \\
lst & MODIS MYD11A1 (Aqua)\textsuperscript{*} & 1\,km & daily \\
dew, rh, wind\_speed, wind\_dir, pressure & ERA5-Land hourly & 9\,km & daily \\
precipitation & CHIRPS & 5\,km & daily \\
refl\_i1, refl\_i2, refl\_m11 & VIIRS VNP09GA (C2) & 500\,m & daily \\
fire (target) & VIIRS SNPP + NOAA-20 C2 & 375\,m & daily \\
\bottomrule
\end{tabular}
\caption{Input channels and their source products. \textsuperscript{*}When Aqua does not see the ground, LST falls back to ERA5-Land skin temperature, with a mean error of 4.0  \textdegree C against real Aqua LST, compared with 11.2 \textdegree C for a daily mean.}
\label{tab:channels}
\end{table*}

\subsection{Dataset splits}
\label{app:sec:splits}

The dataset splits are stratified by region, so the three seeds change which events enter each partition while keeping the regional composition approximately fixed (\cref{tab:splits}). Differences across seeds therefore mainly reflect test-set composition. Persistence follows the same overall trend as model AP (\cref{tab:afseq_seeds}), although not perfectly, indicating that it captures much, but not all, of the variation in test difficulty.

\paragraph{Why the splits are by event and not by year.}
The year distribution is also similar across train, validation, and test despite not being explicitly stratified (\cref{tab:splits}). We do not use temporal folds because AFSeq currently covers only four calendar years, with incomplete coverage at the beginning and end of the collection period. More importantly, year and region are not independent: some years contain substantially different regional compositions, and one year contains no U.S. events. A temporal split would therefore vary both year and geography simultaneously, making it difficult to determine which factor explains a performance change. We instead use event-level splits for the main benchmark and evaluate spatial generalization separately by holding out entire regions (see \cref{app:sec:loro}). Generalization to unseen years remains an important direction once AFSeq covers more complete seasons.

\begin{table*}[t]
\centering
\setlength{\tabcolsep}{4pt}
\begin{tabular}{llrrrrrrrr}
\toprule
& & & \multicolumn{3}{c}{Region} & \multicolumn{4}{c}{Start year} \\
\cmidrule(lr){4-6}\cmidrule(lr){7-10}
Seed & Split & $n$ & S.\,Am. & S.\,Eur. & U.S. & 2023 & 2024 & 2025 & 2026 \\
\midrule
5 & Train & 835 & 317 & 223 & 295 & 89 & 336 & 357 & 53 \\
 & Val & 179 & 68 & 48 & 63 & 10 & 80 & 79 & 10 \\
 & Test & 180 & 68 & 48 & 64 & 17 & 77 & 73 & 13 \\
\midrule
7 & Train & 835 & 317 & 223 & 295 & 77 & 351 & 356 & 51 \\
 & Val & 179 & 68 & 48 & 63 & 18 & 71 & 77 & 13 \\
 & Test & 180 & 68 & 48 & 64 & 21 & 71 & 76 & 12 \\
\midrule
23 & Train & 835 & 317 & 223 & 295 & 82 & 350 & 349 & 54 \\
 & Val & 179 & 68 & 48 & 63 & 18 & 67 & 81 & 13 \\
 & Test & 180 & 68 & 48 & 64 & 16 & 76 & 79 & 9 \\
\bottomrule
\end{tabular}
\caption{Split composition per seed, by region and by start year. Region counts are identical across seeds by construction because the splits are stratified by region. The year balance instead results from random sampling and is not explicitly constrained.}
\label{tab:splits}
\end{table*}

\subsection{Regional coverage}
\label{app:sec:coverage}

\cref{tab:regions} summarizes the three regions represented in AFSeq. South American events tend to last longer and contain larger active-fire areas, while S. Europe is dominated by shorter and smaller events. The United States shows the widest range of event sizes and includes the largest fires in the benchmark. Persistence is lowest in S. Europe, suggesting greater day-to-day changes in the fire mask and making prediction from the previous observation alone more challenging.

\begin{table*}[t]
\centering
\setlength{\tabcolsep}{4pt}
\small
\begin{tabular}{lrrrrrr}
\toprule
Region & Events & Area (ha) & Fire days & Fire px & Persistence \\
 & & median  & median (max) & median & AP \\
\midrule
South America & 453 & 1,088 & 10 (49) & 7,017 & 0.128 \\
S. Europe & 319 & 625 & 2 (25) & 1,969 & 0.095 \\
United States & 422 & 1,327 & 5 (42) & 4,227 & 0.122 \\
\bottomrule
\end{tabular}
\caption{Per-region coverage of AFSeq. \emph{Fire days} counts only days with active detections; \emph{fire px} is the median number of active-fire pixels per event. Persistence is the AP at $t{+}1$ of copying the day $t$ mask, measured over every event of the region.}
\label{tab:regions}
\end{table*}

\subsection{Regional characteristics}
\label{app:sec:region_analysis}

\begin{figure}[t]
\centering
\includegraphics[width=\linewidth]{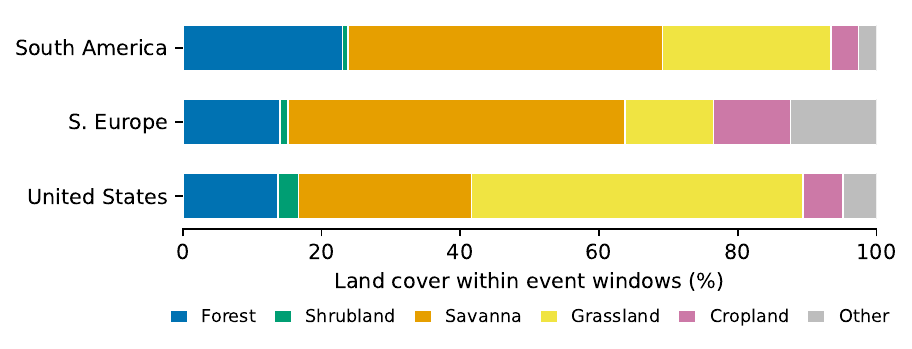}
\caption{IGBP land cover within the event windows of each region, aggregated into fuel types.}
\label{fig:fuels}
\end{figure}

\begin{figure*}[t]
\centering
\includegraphics[width=\linewidth]{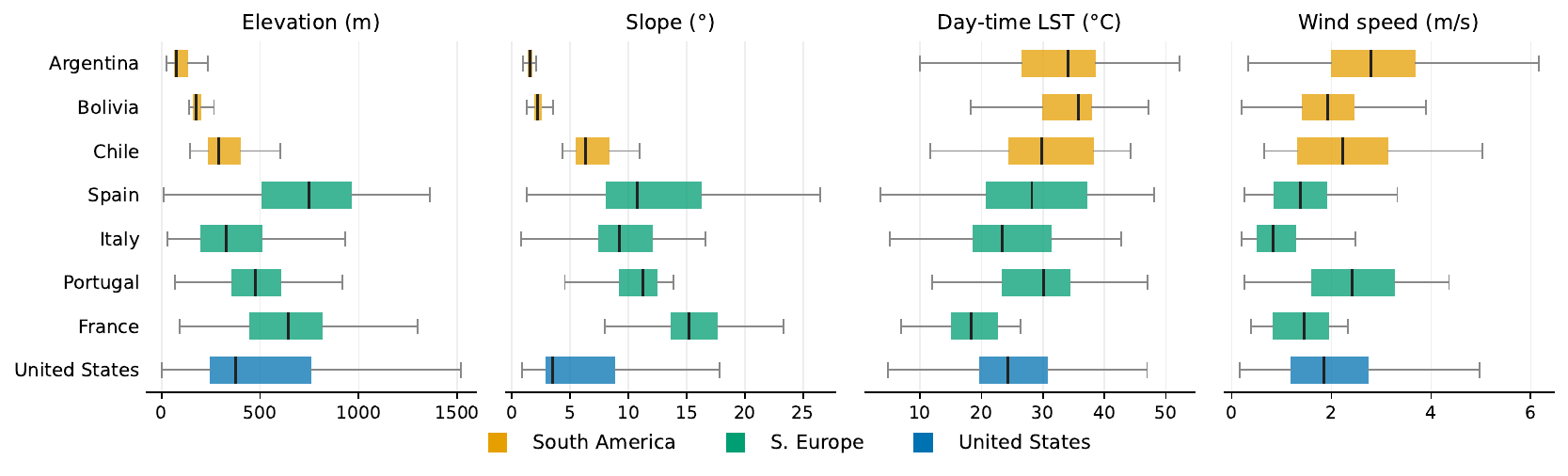}
\caption{Terrain and weather conditions on the day of the fire by country. Boxes show the central 50\% of the observations, with the median indicated by the line inside each box.}
\label{fig:countrydist}
\end{figure*}

The three regions also differ in land cover, terrain, and weather conditions. \cref{fig:fuels} shows the land-cover distribution across regions, while \cref{fig:countrydist} provides country-level differences in terrain and day-of-fire weather. This distinction is important because regional averages can hide substantial variation within each region. For example, South America includes Argentina, Bolivia, and Chile, which differ considerably in their environmental conditions.

Argentina is dominated by savanna and relatively flat, low-elevation terrain. Bolivia is the only part of the benchmark where forest is the dominant land-cover type, while Chile has steeper and higher terrain. Within S. Europe, France contains the steepest terrain in the benchmark and the lowest daytime LST, whereas Portugal shows the highest wind speeds. The United States covers a broad range of terrain, including both plains and mountainous areas.

These differences help explain the cross-region results in \cref{app:sec:loro}. Generalization across regions is affected not only by the number of available events, but also by differences in land cover and terrain.

\subsection{Comparison with WSTS+}
\label{app:sec:region_analysis}

\begin{figure*}[t]
\centering
\includegraphics[width=\linewidth]{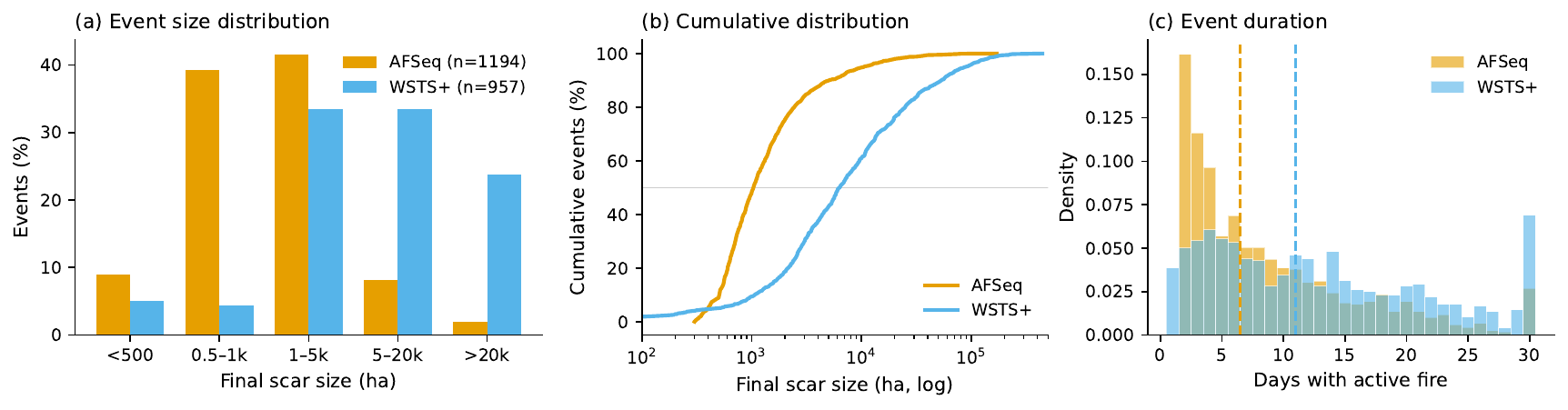}
\caption{Comparison between AFSeq and WSTS+. (a) Final fire size by range; (b) cumulative distribution on a log scale; (c) number of days with active fire.}
\label{fig:dscmp}
\end{figure*}

\cref{fig:dscmp} compares the event-size distribution of AFSeq with WSTS+. The two benchmarks cover substantially different fire sizes. AFSeq is concentrated on smaller events, particularly between 500 and 5,000 ha, whereas WSTS+ contains a much larger proportion of events above 5,000 ha. AFSeq events are also shorter on average.

\section{UTAE-AR Details}
\label{app:sec:utae_det}

\subsection{UTAE-AR Architecture}
\label{sec:supp_utaear}

UTAE-AR combines a U-TAE encoder with a multi-scale ConvLSTM decoder that predicts the forecast sequence autoregressively. The input history $x_{1:T}$ is encoded once. U-TAE produces features at three spatial resolutions, while L-TAE summarizes the temporal information at the bottleneck and provides attention weights for the skip connections. During forecasting, a ConvLSTM at each resolution updates these features over time, and the resulting representations are decoded through the same U-TAE upsampling blocks. The model has 3.16 M parameters, substantially fewer than the Res18-UTAE baseline.

\subsection{What differs from UTAE at t{+}1}

At $t{+}1$, UTAE-AR has not yet fed back any predicted mask, so its first forecast relies only on the observed history, as in UTAE. The two models still do not follow the same path. In UTAE-AR, the L-TAE output initializes a ConvLSTM cell at each level, and this cell performs one update with the encoding of $x^{(t)}$ before the decoder produces the mask, as described in Eqs. 1 to 4 of the main paper. On top of this recurrent step, UTAE-AR has more parameters than UTAE, 3.16 M against 0.89 M, and during training it is supervised over the next five days instead of only the next one.

The larger capacity alone does not explain the gain, since Res18-UTAE has 4.7 times more parameters than UTAE-AR and still obtains a lower AP on both AFSeq and WSTS+, as shown in Tables 2 and 3 of the main paper. The difference therefore comes mainly from the autoregressive design: supervising several future days encourages the encoder to learn a representation that supports fire propagation beyond a single next-day prediction, and the recurrent step carries this representation into the first forecast. As a result, UTAE-AR obtains the best $t{+}1$ performance across all seeds (\cref{tab:afseq_seeds}).


\section{Experimental details}
\label{app:sec:exp_det}

\subsection{Training}
\label{sec:supp_training}

\paragraph{AFSeq models.}
All models evaluated on AFSeq use the same training setup, so differences between rows reflect only the model architecture. We train for 60 epochs with Adam, using a learning rate of $10^{-3}$, weight decay of $10^{-5}$, batch size 4, and a BCE-Dice loss with positive-class weight 20. The learning rate is reduced by a factor of 0.5 when validation AP does not improve for three epochs. Inputs are $128\times128$ crops centered on the fire region at day $t$, with random spatial jitter. Each sample contains up to 5 days of history and up to 5 future days. Shorter histories and forecast sequences are zero-padded and masked so that padded days do not contribute to training or evaluation. Checkpoints are selected using validation AP. Results are averaged over three splits with seeds 5, 7, and 23.

\paragraph{WSTS+ baselines.}
For the WSTS+ experiments reported in Tables 3 and 4 of the main paper, we retain the training configuration used by the original authors for all models to ensure comparability with their reported results. This includes a ResNet-18 encoder initialized with ImageNet weights, focal loss with $\gamma=2$, AdamW with a learning rate of $10^{-3}$, 10{,}000 training steps, a batch size of 64, 32-bit precision, and selection of the checkpoint with the highest validation AP.

\paragraph{Compute environment.}
Experiments are run on NVIDIA A100-SXM4-80GB GPUs using Python 3.11.16, PyTorch 2.13 with CUDA 13.0, PyTorch Lightning 2.6.5, segmentation-models-pytorch 0.5.0, and NumPy 2.4.6. Training a single AFSeq model takes about 3-6 GPU hours.

\subsection{Deployment}
\label{app:subsec:supp_deploy}

\cref{tab:inference} reports inference cost on a single NVIDIA A100 with batch size~1 and the input configuration used in our experiments ($T{=}5$ days, $128\times128$\,px). UTAE-AR has 3.16 M parameters, about 4.7$\times$ fewer than Res18-UTAE.

\begin{table}[t]
\centering
\setlength{\tabcolsep}{4pt}
\resizebox{\columnwidth}{!}{%
\begin{tabular}{lrrrrrr}
\toprule
Model & Params (M) & Predicted days & GFLOPs & Latency (ms) & ms / day & Peak mem.\ (MiB) \\
\midrule
UTAE-AR (ours) & 3.16 & 5 & 124.5 & 21.3 & \textbf{4.3} & 139 \\
Res18-UTAE     & 14.7 & 1 & 11.4 & 14.5 & 14.5 & 120 \\
UTAE           & 0.89 & 1 & 38.8 & 8.2 & 8.2 & 210 \\
\bottomrule
\end{tabular}}
\caption{Inference cost per event on one A100 with batch size 1. UTAE-AR predicts the full five-day horizon in each forward pass, while Res18-UTAE and UTAE predict only $t{+}1$. UTAE-AR has 4.7$\times$ fewer parameters than Res18-UTAE, although it requires more FLOPs.}
\label{tab:inference}
\end{table}

\subsection{Fusion details}
\label{sec:supp_fusion}

To train a single model on both benchmarks, we represent them using the same input channels. We use the seven-channel \emph{Veg} configuration from WSTS+: \texttt{m11}, \texttt{i2}, \texttt{i1}, \texttt{ndvi}, \texttt{evi2}, fire detection hour, and the binary fire mask. For WSTS+, these channels are directly selected from its processed inputs. AFSeq provides the same information natively: M11 is obtained from VNP09GA together with I1 and I2, while the detection hour travels as a second band of the daily active-fire raster. No channel is replaced by a constant value.

The original train, validation, and test partitions of both datasets are preserved during fusion. For each fold $f$, we define
$\mathrm{train}(f) = \mathrm{WSTS{+}}*{\mathrm{train}}(f) \cup
\mathrm{AFSeq}*{\mathrm{train}}$,
and construct the validation and test sets in the same way. Thus, events assigned to the test set in either benchmark are never used for training. Three fires appearing in both datasets are removed from the WSTS+ side to avoid duplicate events, while the original AFSeq splits remain unchanged.

Models are trained on the combined training set but evaluated separately on the original test set of each benchmark.


\section{Supplementary Results}
\label{app:sec:supp_results}

\subsection{WSTS+ per-fold AP}
\label{sec:supp_folds}

\cref{tab:wsts_folds} reports the WSTS+ results for each fold. Performance varies substantially across folds, from about 0.31 to 0.55 AP, but the pattern is consistent across all three feature sets. In every case, the folds follow the same ordering as the persistence baseline (fold~1 $<$ fold~0 $<$ fold~2 $<$ fold~3), with a strong correlation between persistence and UTAE-AR performance. This suggests that part of the variation across folds reflects differences in their intrinsic predictability. The cross-fold mean therefore hides substantial variation in difficulty across the dataset.

\begin{table}[t]
\centering
\setlength{\tabcolsep}{2.5pt}
\scriptsize
\begin{tabular}{lrrrrr}
\toprule
 & fold 0 & fold 1 & fold 2 & fold 3 & mean $\pm$ std \\
\midrule
Persistence & 0.134 & 0.067 & 0.182 & 0.233 & 0.154 $\pm$ 0.071 \\
\midrule
UTAE-AR, Veg   & 0.378 & 0.309 & 0.495 & 0.539 & 0.429 $\pm$ 0.105 \\
UTAE-AR, Multi & 0.410 & 0.318 & 0.508 & 0.549 & \textbf{0.445} $\pm$ 0.104 \\
UTAE-AR, All   & 0.406 & 0.311 & 0.501 & 0.550 & 0.441 $\pm$ 0.106 \\
\bottomrule
\end{tabular}
\caption{UTAE-AR on WSTS+ per fold, for the three feature sets of
\cref{tab:wstsplus}. Test AP at $t{+}1$ on the 128-pixel crop, with five days
of history as in the published protocol. The fold ordering is identical across
feature sets and follows the persistence floor ($r = +0.98$).}
\label{tab:wsts_folds}
\end{table}

\subsection{AFSeq per-partition AP}
\label{sec:supp_seedap}

\cref{tab:afseq_seeds} reports the per-partition results for the main benchmark. As in \cref{sec:supp_folds}, model performance varies consistently with the persistence baseline across seeds. UTAE-AR achieves the best AP for every seed and also shows the lowest variation across them.

\begin{table*}[t]
\centering
\setlength{\tabcolsep}{4pt}
\small
\begin{tabular}{lcccc}
\toprule
Model & partition 5 & partition 7 & partition 23 & mean $\pm$ std \\
\midrule
Persistence & 0.120 & 0.114 & 0.139 & 0.124 $\pm$ 0.011 \\
\midrule
U-Net & 0.216 & 0.232 & 0.278 & 0.242 $\pm$ 0.032 \\
ConvNeXt & 0.215 & 0.258 & 0.289 & 0.254 $\pm$ 0.037 \\
ConvLSTM & 0.273 & 0.305 & 0.328 & 0.302 $\pm$ 0.028 \\
UTAE & 0.268 & 0.290 & 0.336 & 0.298 $\pm$ 0.035 \\
Res18-UTAE & 0.309 & 0.320 & 0.335 & 0.321 $\pm$ 0.013 \\
UTAE-AR (ours) & 0.368 & 0.371 & 0.407 & 0.382 $\pm$ 0.022 \\
\bottomrule
\end{tabular}
\caption{AFSeq test AP per partition, pooled at $t{+}1$.}
\label{tab:afseq_seeds}
\end{table*}

\subsection{AP per wildfire size}
\label{sec:supp_apsize}

\begin{figure*}[t]
\centering
\includegraphics[width=\linewidth]{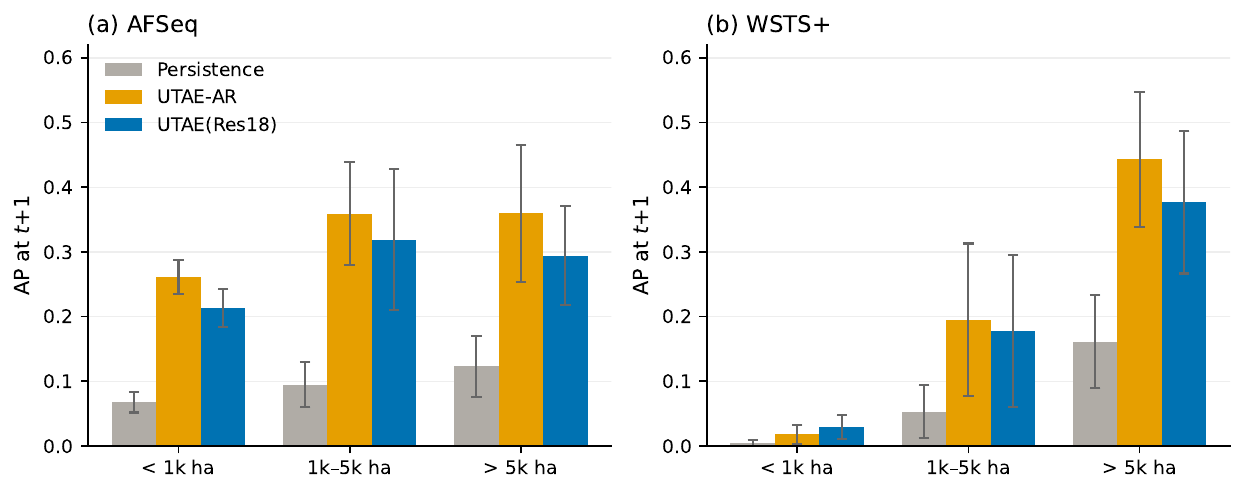}
\caption{Next-day AP by fire size in AFSeq and WSTS+. Fires are grouped by final burned area into $<$1k ha, 1k-5k ha, and $>$5k ha. Error bars show the standard deviation across three splits-seeds for AFSeq and four folds for WSTS+.}
\label{fig:apsize}
\end{figure*}

\cref{fig:apsize} compares next-day AP across three fire-size groups in AFSeq and WSTS+. The two benchmarks show different relationships between fire size and prediction performance.

In AFSeq, performance improves substantially from fires below 1k ha to the 1k-5k ha group, and then remains similar for UTAE-AR while decreasing slightly for Res18-UTAE. In contrast, WSTS+ shows a much stronger dependence on fire size: performance is very low for fires below 1k ha and increases consistently toward the largest group. The persistence baseline follows the same general trend, particularly in WSTS+.

\subsection{WSTS 12-fold evaluation}
\label{sec:wsts12}

The four-fold WSTS+ protocol (\cref{tab:wsts_folds}) differs from the original 12-fold protocol used by WildfireSpreadTS. For completeness, we also evaluate UTAE-AR under this setting, using one year for testing, one for validation, and the remaining two for training. \cref{tab:wsts12} reports the resulting $t{+}1$ AP together with the persistence baseline for each held-out year.

The results are consistent with those observed under the four-fold protocol and on AFSeq: performance varies strongly across test years and closely follows the persistence baseline. In contrast, changing only the validation year produces much smaller differences. This suggests that most of the variation across the twelve folds comes from differences between test years rather than from the choice of validation split.

Therefore, the twelve folds largely provide repeated evaluations of the same four held-out years. The four-fold protocol captures this variation more directly while avoiding multiple closely related measurements of each test year.

\begin{table*}[t]
\centering
\setlength{\tabcolsep}{4pt}
\begin{tabular}{crrrrr}
\toprule
Fold & Train years & Val & Test & Persistence & UTAE-AR \\
\midrule
0 & 2020, 2021 & 2019 & 2018 & 0.192 & 0.511 \\
1 & 2019, 2021 & 2020 & 2018 & 0.192 & 0.495 \\
2 & 2019, 2020 & 2021 & 2018 & 0.192 & 0.485 \\
3 & 2020, 2021 & 2018 & 2019 & 0.112 & 0.374 \\
4 & 2018, 2021 & 2020 & 2019 & 0.112 & 0.357 \\
5 & 2018, 2020 & 2021 & 2019 & 0.112 & 0.356 \\
6 & 2019, 2021 & 2018 & 2020 & 0.193 & 0.475 \\
7 & 2018, 2021 & 2019 & 2020 & 0.193 & 0.497 \\
8 & 2018, 2019 & 2021 & 2020 & 0.193 & 0.483 \\
9 & 2019, 2020 & 2018 & 2021 & 0.276 & 0.561 \\
10 & 2018, 2020 & 2019 & 2021 & 0.276 & 0.579 \\
11 & 2018, 2019 & 2020 & 2021 & 0.276 & 0.562 \\
\midrule
\multicolumn{4}{l}{Mean $\pm$ std over 12 folds} & 0.193 $\pm$ 0.058 & 0.478 $\pm$ 0.074 \\
\bottomrule
\end{tabular}
\caption{UTAE-AR on the original WSTS 12-fold protocol. Persistence is the
copy-the-day-$t$-mask floor of the held-out year; folds sharing a test year
share it by construction. AP is pooled at $t{+}1$ on the full test year.}
\label{tab:wsts12}
\end{table*}

\subsection{Res18-UTAE as a Seq-to-Seq baseline}
To test whether the multi-day gain of UTAE-AR comes only from training on several days, we train on \dsname{} a Res18-UTAE with $F=5$ segmentation heads, one per forecast day. All heads share the same encoder and decoder, predict their day directly from the observed history, and are trained with the sum of the per-day losses. \textbf{At $t{+}1$, this model reaches 0.229 AP, against 0.321 for the single-head Res18-UTAE and 0.382 for UTAE-AR. It also stays below the recursive Res18-UTAE at every later horizon.} Supervising several days therefore does not explain the multi-day gain of UTAE-AR, which comes from its autoregressive design.


\section{Leave-one-region-out on \dsname{}}
\label{app:sec:loro}

The standard AFSeq splits are stratified by region, so the main benchmark evaluates fires from the same three regions observed during training. To test cross-region generalization, we additionally perform leave-one-region-out (LORO) evaluation. For each region, models are trained and validated on the other two and evaluated on the held-out region. \cref{tab:loro} compares this result with the persistence baseline and with the corresponding performance under the standard benchmark split. Since baseline difficulty differs across regions, we focus on changes within each region rather than comparing absolute AP across regions.

The largest degradation occurs when South America is excluded from training, where both architectures lose a substantial amount of performance. In contrast, performance does not decrease when S. Europe is held out. The United States shows a clearer difference between architectures: UTAE-AR retains most of its standard-split performance, whereas Res18-UTAE shows a larger drop.

These results also show that regional difficulty and cross-region transferability are not necessarily the same. S. Europe has the lowest persistence baseline but transfers well from the other regions, while South America has a higher persistence baseline and shows the largest loss when excluded from training. This suggests that difficulty within a region does not directly determine how well models generalize to it from other regions.

\begin{table*}[t]
\centering
\setlength{\tabcolsep}{4pt}
\small
\begin{tabular}{llrrr}
\toprule
Held out & Model & LORO & Standard split & $\Delta$ \\
\midrule
\multirow{3}{*}{South America} & Persistence       & 0.128 & --                  & -- \\
              & UTAE-AR (ours)    & \textbf{0.206} & 0.364 $\pm$ 0.051 & $-0.158$ \\
              & Res18-UTAE       & 0.181 & 0.322 $\pm$ 0.091 & $-0.141$ \\
\midrule
\multirow{3}{*}{S.\ Europe}    & Persistence       & 0.095 & --                  & -- \\
              & UTAE-AR (ours)    & \textbf{0.243} & 0.279 $\pm$ 0.053 & $-0.036$ \\
              & Res18-UTAE       & 0.193 & 0.202 $\pm$ 0.017 & $-0.009$ \\
\midrule
\multirow{3}{*}{United States} & Persistence       & 0.122 & --                  & -- \\
              & UTAE-AR (ours)    & \textbf{0.305} & 0.343 $\pm$ 0.075 & $-0.038$ \\
              & Res18-UTAE       & 0.198 & 0.264 $\pm$ 0.087 & $-0.066$ \\
\bottomrule
\end{tabular}
\caption{Leave-one-region-out (LORO) evaluation on AFSeq. LORO models are trained on
two regions and evaluated on the held-out region. \emph{Standard split} reports
performance when the held-out region is represented during training, and
$\Delta$ is the difference between LORO and standard-split AP.}
\label{tab:loro}
\end{table*}


\section{Variable Analysis}
\subsection{Input importance across the forecast horizon}
\label{sec:occlusion}

\begin{figure*}[t]
\centering
\includegraphics[width=\linewidth]{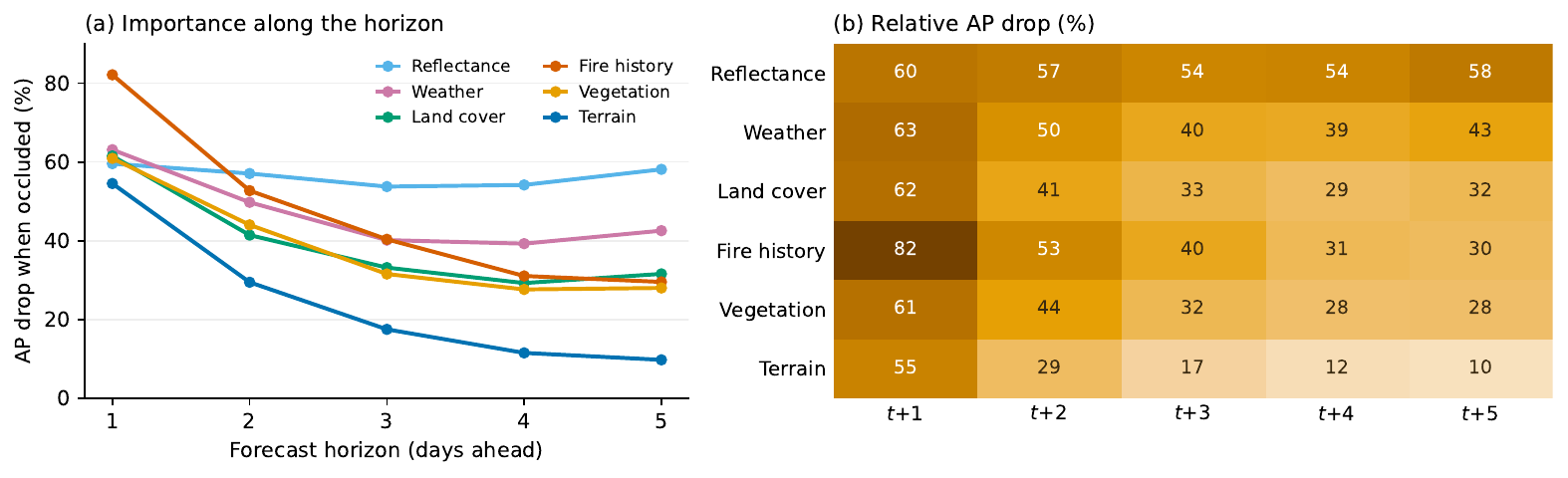}
\caption{Input importance across the forecast horizon. (a) Relative AP drop when each input group is replaced by its dataset mean. (b) The same results shown across horizons, with input groups ordered by their importance at $t{+}5$.}
\label{fig:occlusion}
\end{figure*}

Input importance is usually evaluated only for next-day prediction, but autoregressive forecasting allows it to change with the prediction horizon. To study this effect, we occlude one group of input channels at a time by replacing them with their dataset mean, both in the input history and in the available future covariates. We then measure the resulting drop in AP at each horizon. \cref{fig:occlusion} reports the results for UTAE-AR on AFSeq.

The relative importance of several inputs changes substantially over the rollout. Fire history is most important for the next-day prediction, but its effect decreases at longer horizons. This is expected because later predictions increasingly depend on fire states generated by the model itself rather than directly on the observed mask. In contrast, reflectance remains important throughout the full horizon and becomes the most influential input group from $t{+}2$ onward. Weather shows a similar, although weaker, pattern.

Static inputs decrease more clearly in importance with lead time. Terrain, land cover, and vegetation have a larger effect on the first predictions and a smaller effect toward the end of the rollout. These variables describe persistent properties of the landscape, but do not directly capture the evolving state of the fire. Overall, the results suggest that the information used by the model changes with the forecasting horizon: the observed fire state is particularly important for next-day prediction, whereas spectral and weather information remain more relevant for longer rollouts.

The magnitude of the AP drop should not be compared directly across input groups of different sizes, since occluding a larger number of channels produces a larger perturbation. We therefore focus primarily on how the importance of each input group changes across prediction horizons.

\subsection{Future covariates ablation}
\label{sec:supp_fcabl}

Multi-day forecasting requires future weather covariates. We compare three settings: \emph{ground-truth future covariates}, which provide an upper bound; \emph{GFS forecast}, which represents the information available at deployment; and \emph{persisted} covariates, which keep the day-$t$ values fixed throughout the rollout. The history is always observed, and only future covariates change.

\cref{tab:fcabl} evaluates both the model used in the main experiments, trained using the ground-truth weather available in the dataset, and a model trained directly with GFS forecasts. For the first model, using GFS forecasts at inference retains most of the performance obtained with ground-truth future weather and substantially outperforms repeated day-$t$ weather at longer horizons.

\begin{table*}[t]
\centering
\setlength{\tabcolsep}{4pt}
\small
\begin{tabular}{llrrrrr}
\toprule
Trained on & Future covariates at inference & $t{+}1$ & $t{+}2$ & $t{+}3$ & $t{+}4$ & $t{+}5$ \\
\midrule
\multirow{3}{*}{Ground-truth future covariates}
& Ground-truth future covariates    & 0.368 & 0.318 & 0.274 & 0.236 & 0.222 \\
& GFS forecast issued on day $t$   & 0.368 & 0.268 & 0.236 & 0.198 & 0.192 \\
& Day-$t$ values frozen            & 0.368 & 0.222 & 0.154 & 0.118 & 0.104\\
\midrule
\multirow{3}{*}{GFS forecast}
& GFS forecast issued on day $t$   & 0.360 & 0.286 & 0.243 & 0.200 & 0.181 \\
& Ground-truth future covariates              & 0.360 & 0.281 & 0.229 & 0.187 & 0.168 \\
& Day-$t$ values frozen            & 0.360 & 0.192 & 0.126 & 0.090 & 0.076 \\
\bottomrule
\end{tabular}
\caption{AP across forecast horizons under different future-covariate settings on AFSeq (seed 5). The history is always observed and only future covariates change.}
\label{tab:fcabl}
\end{table*}


\section{Additional qualitative results}
\label{sec:add_results}

\begin{figure*}[t]
\centering
\includegraphics[width=\linewidth]{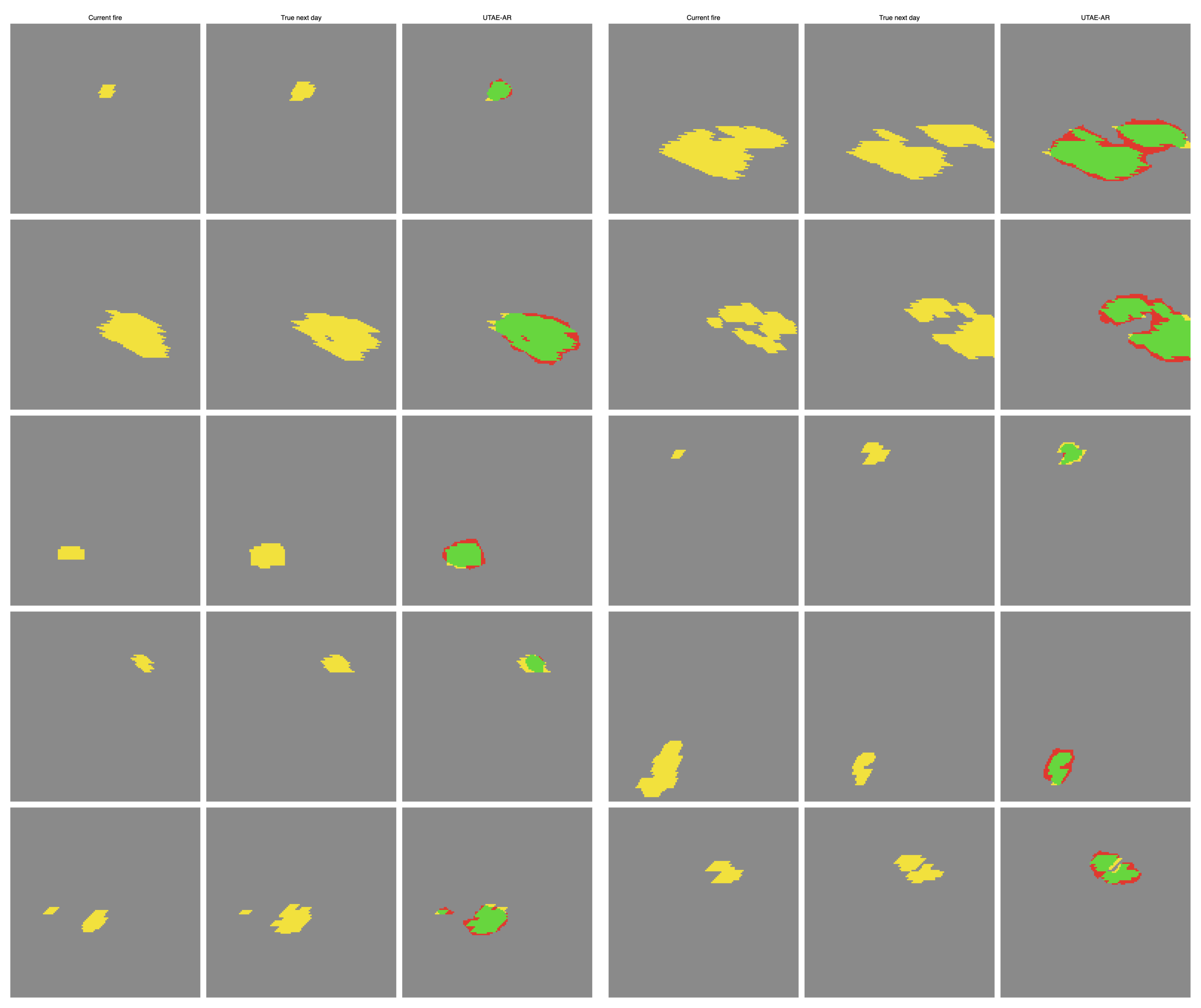}
\caption{The ten highest-AP UTAE-AR predictions at $t{+}1$ on the AFSeq test set, ordered from left to right and top to bottom. Green denotes true positives, red false positives, and yellow false negatives.}
\label{fig:qual_best}
\end{figure*}

\begin{figure*}[t]
\centering
\includegraphics[width=\linewidth]{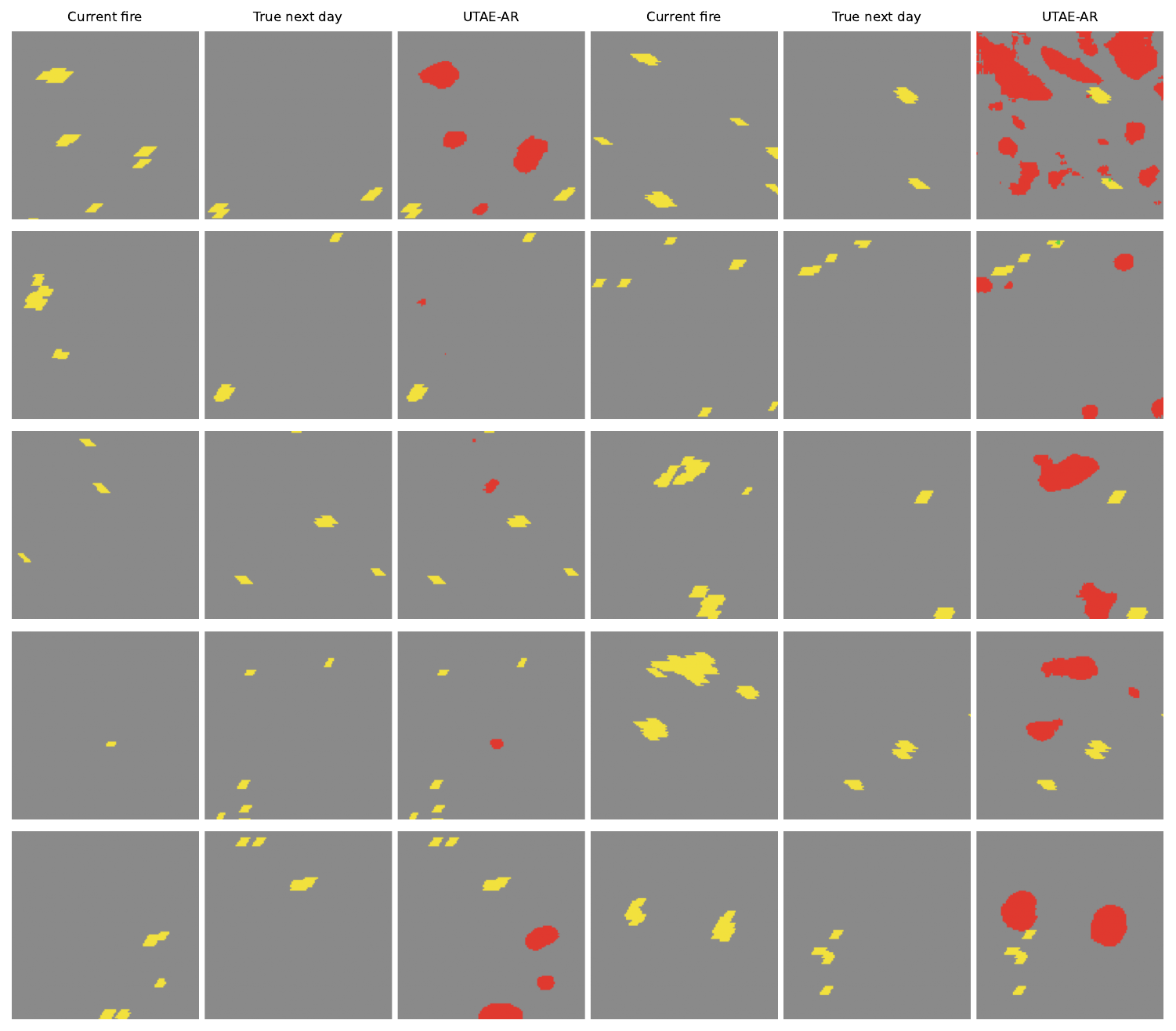}
\caption{The ten lowest-AP UTAE-AR predictions at $t{+}1$ on the AFSeq test set, ordered from left to right and top to bottom. Green denotes true positives, red false positives, and yellow false negatives. These cases often correspond to fragmented fires with activity distributed across disconnected regions.}
\label{fig:qual_worst}
\end{figure*}

\cref{fig:qual_best} and \cref{fig:qual_worst} show the ten best and worst UTAE-AR predictions at $t{+}1$ on the AFSeq test set (seed 23).

The best predictions correspond mainly to compact, contiguous fire fronts, where the model captures the overall direction of spread with limited overprediction. The worst cases are more fragmented, with disconnected or newly active patches that are harder to infer from the previous fire state. These examples suggest that most qualitative failures occur when fire activity changes abruptly across space rather than along a continuous front.

\end{document}